# Procrastination Is All You Need: Exponent Indexed Accumulators for Floating Point, Posits and Logarithmic Numbers

Vincenzo Liguori        Ocean Logic Pty Ltd        enzo@ocean-logic.com

**Abstract**
**This paper discusses a simple and effective method for the summation of long sequences of floating point numbers. The method comprises two phases: an accumulation phase where the mantissas of the floating point numbers are added to accumulators indexed by the exponents and a reconstruction phase where the actual summation result is finalised. Various architectural details are given for both FPGAs and ASICs including fusing the operation with a multiplier, creating efficient MACs. Some results are presented for FPGAs, including a tensor core capable of multiplying and accumulating two 4x4 matrices of bfloat16 values every clock cycle using ~6,400 LUTs + 64 DSP48 in AMD FPGAs at 700+ MHz. The method is then extended to posits and logarithmic numbers.**

## 1. Introduction

The summation of long sequences of floating point numbers is an extremely important operation, integral part of dot products and matrix multiplications. These, in turn, are ubiquitous in computational science as well as Convolutional Neural Networks, Large Language Models and much more: an efficient implementation clearly matters.

The basics about the method described here is in section 2. This section also includes some general implementation details and possible optimizations. Fusion with the multiply operation is also discussed.

More implementation details are given in sections 3 and 4, dedicated to FPGAs and ASICs respectively.

Sections 5 and 6 extend the basic concept to posits and logarithmic numbers.

## 2. Floating Point

The objective is to add a series of N floating point numbers in the form $m2^e$. As anticipated, the method consists in two phases. First, all the N numbers are processed by creating partial sums of all the mantissas that have the same exponent. In the next step this procrastination is finally resolved with the final result being reconstructed from said partial sums. See Eq. 1.

$$\sum_{i=0}^{N-1} m_i 2^{e_i} = \sum_{i=0}^{N_e-1} S_i 2^i \text{ where } S_i = \sum_{\forall k : e_k = i} m_k \quad \text{Eq.1}$$

Where $N_e = 2^{ne}$ with ne being the number of bits of the exponent. The mantissa are signed integers. Eq. 1 simply groups all the values with the same exponent that is then factored out. See the pseudo-code below for the first step:

```
// Accumulation phase
m[N]; // Signed mantissas of the N floating point numbers
e[N]; // Exponents of the N floating point numbers
S[Ne]; // Partial sums, initialised to 0
for(i=0;i<N;i++)
```

```
    S[e[i]] += m[i]; // Create the partial sums
```

After the accumulation phase is completed, the reconstruction of the final result can be done by shifting and adding the partial sums, according to Eq. 1. However, using Eq.1 directly will require dealing with very large numbers. A much simpler method is to use a small accumulator, shifting out the LSB of the result one at the time, as they are calculated, as shown in the pseudo-code below:

```
// Reconstruction phase
a = 0; // Initialise the reconstruction accumulator
for(i=0;i<Ne;i++){ // For each exponent value
  a += S[i];
  S[i] = 0; // Clear for the next operation
  Output a & 1; // Output one bit of the final result, from LSB
  a >>= 1;
}
Output a; // Top bits of the final result
```

Note that:

- For many cases of interest, N is in the order of tens of thousand, thus the reconstructing phase is negligible in comparison to the accumulation phase.

- The reconstructing accumulator (and adder) only needs to be one bit bigger than the partial sums: no need to deal with large numbers, large additions or shifts.

- All the bits shifted out by the pseudo-code above, together with the final value of the accumulator, represent the **exact result** of the sum of all the N floating point numbers. It follows that we can control the precision of the result by discarding some of the bits shifted out.

- If, during the accumulation phase, we keep track of the maximum ($max_e$) and minimum ($min_e$) exponents encountered, then, during the reconstruction phase, we only need to loop between $min_e$ and $max_e$ (instead of from 0 to $N_e$-1) because all the other partial sums will be zero. The result, in this case, will be up to a scaling factor of $2^{min_e}$.

- More reconstruction strategies are discussed in section 2.3..

Next we look at hardware implementation.

## 2.1. Hardware Implementation

Fig.1 shows a possible hardware implementation of the algorithm above.

As before, ne is the number of bits of the exponent, $N_e = 2^{ne}$, nm the number of bits of the mantissa. Each partial sum register contains an extra nv bits to avoid overflowing. Absolutely worst case nv = ceil($\log_2$(N)). This is extremely pessimistic as it assumes that all the N numbers being added have the same exponent and the largest possible mantissa.

The circuit starts with all the partial sum registers set to zero. Then, during the accumulation phase, for every floating point number that is input, the exponent is used to read the corresponding partial sum. The reconstructed mantissa (hidden bit included) is then added or subtracted to it (according to the sign) and the result is stored back in the same register. This happens in the same clock cycle. Not shown in the figure is the logic that detects a floating point value of zero and disables writing back the value.

During the reconstruction phase, the partial sums are read sequentially from the registers (for example, by providing sequential addresses through the Exponent input) and the final result is reconstructed with shifts and adds, as shown in the pseudo-code. During this phase, we can also take the opportunity to set to zero the registers as they are read out. This is not shown in the picture for clarity. We do not need to clear the partial sum registers if we want to add more numbers.

The reconstruction phase will take $N_e$ cycles, unless, as explained above, we keep track of $min_e$ and $max_e$. In this case, only $max_e - min_e + 1$ cycles will be required. During the reconstruction, one bit of the final result will be output for each clock cycle, from LSB onwards. At the end, the result register will contain the top bits of the final result.

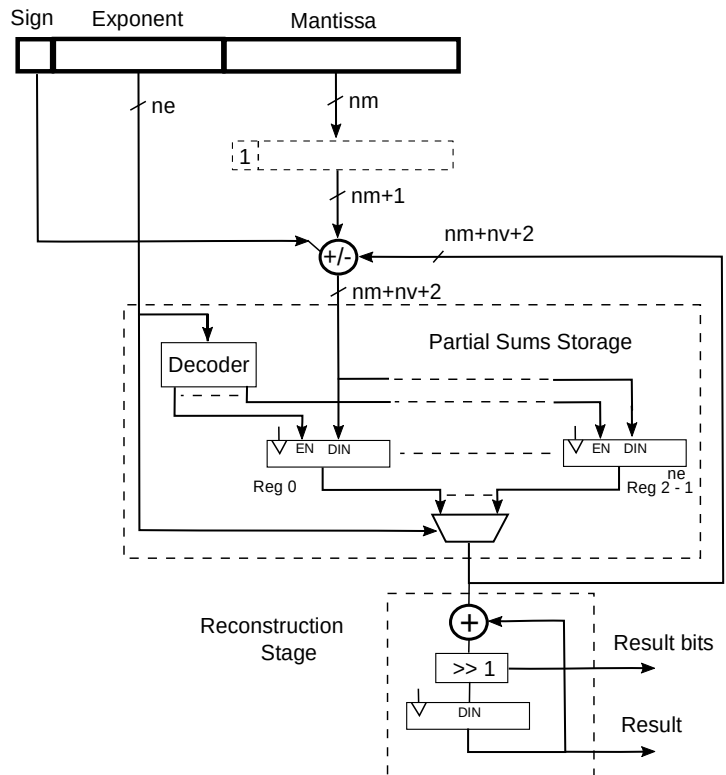

Figure 1 Floating point accumulator architecture.

Note that this design can be easily pipelined. Also note that, for clarity and simplicity, the reconstruction stage is shown in Fig.1 as distinct from the accumulation stage. In reality, since the two adders shown are not used at the same time, there are quite a few ways to share a single adder between the two phases. A similar observation can be made for the result register that can also be shared with the partial sum registers.

This architecture is fast and simple but the large number of registers is a concern. This, as will shall see, is not really a problem for FPGAs, but, nevertheless, it's worth looking at ways to reduce them.

## 2.2. Reducing the Number of Partial Sums Registers

So far we have assigned a partial sum register for each exponent. However, we can reduce the number of such registers by assigning groups of exponents to the same register. Given a parameter k that varies from 0 to ne, we can have $2^{ne-k}$ registers accumulating mantissas that have exponents part of a group of $2^k$ values.

Each mantissa within such group needs to be shifted left by an amount that varies from 0 to $2^k-1$ before being added to a partial sum register. This process is much more difficult to explain in words than showing it in the pseudo-code below.

```
// Accumulation phase
m[N]; // Mantissas of the N floating point numbers
e[N]; // Exponents of the N floating point numbers
S[Ne>>k]; // Partial sums, initialised to 0
maske = (1<<k) -1; // Exponent mask
mask = (1<<(1<<k)) -1; // Output mask
for(i=0;i<N;i++)
  S[e[i]>>k] += m[i]<<(e[i] & maske); // Create the partial sums
// Reconstruction phase
a = 0; // Initialise the reconstruction accumulator
for(i=0;i<(Ne>>k);i++){ // For each group of exponent value
  a += S[i];
  S[i] = 0; // Clear for the next operation
  Output a & mask; // Output 2^k bits of the final result from LSB
  a >>= 1<<k;
}
Output a; // Top bits of the final result
```

So, we now have $2^{ne-k}$ registers. However, we now also need a $[0...2^k-1]$ bit shifter and the partial sums require an additional $2^k$ bits. It's much easier to look at actual hardware on Fig.2.

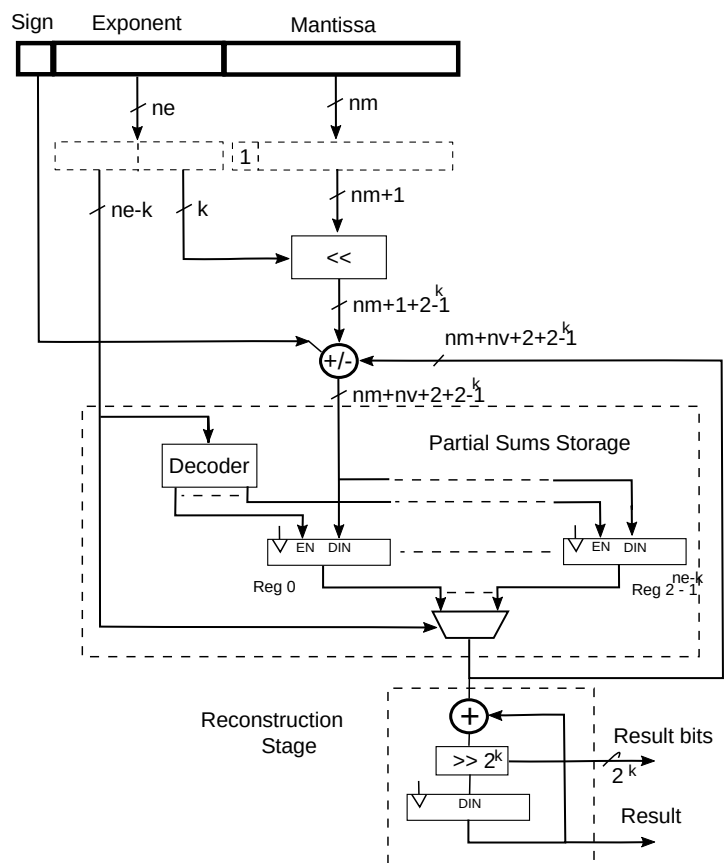

Figure 2 Floating point accumulator with reduced registers.

Essentially, k least significant bits from the exponent are used to left shift the incoming mantissa while the remaining ne – k bits are used to select a partial sum register. As anticipated, the registers require extra $2^k$ bits. Also, during the reconstruction phase $2^k$ bits of the result are shifted out every clock cycle, starting from the LSB.

Note that, for k = 0, we have the architecture described in Fig.1, while, for k = ne, we have the Kulish accumulator [1] (Fig.3).

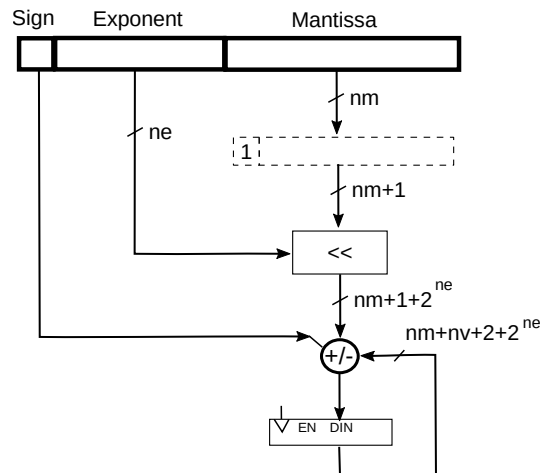

Figure 3 Kulisch accumulator.

## 2.3. Reconstruction Strategies

We have seen that, if we want an exact result, we can go, during the reconstruction phase, from the minimum to the maximum exponent encountered during the accumulation phase.

In most cases the exact result will not be needed and we could conduct the reconstruction using fewer partial sum starting from a few below the maximum exponent encountered to the maximum one. In this case we will get a quicker but less accurate result.

However, we also need to remember that, in order to ensure that the next accumulation operation is correct, all the partial sums in the storage area need to be initialised to zero (i.e. clearing any previous result). If the storage area is implemented as RAM cells (as it's done in FPGAs, see section 3.) it is generally only possible to do this sequentially, one location at the time. In this case, even though we can get a quicker result by using only some partial sums, we still need to clear them all one by one.

In ASIC however, relatively small memories are often implemented as flip-flops and these can be cleared simultaneously with a common synchronous reset line. In this case a quicker and less accurate reconstruction is possible while, at the same time, being quickly ready to start a new operation.

## 2.4. Multiply Accumulator (MAC)

The accumulators described so far can be easily fused with a multiplier to perform multiply/accumulate operations, as show in Fig.4.

Given the two input operands, their mantissas are reconstructed and multiplied together. The exponents are added and the sign of the result is determined by xoring the signs. The resulting sign, exponent and mantissa are then ready to be accumulated as previously described.

If the exponents are biased, as it happens in many floating point formats, we need to take this into account.

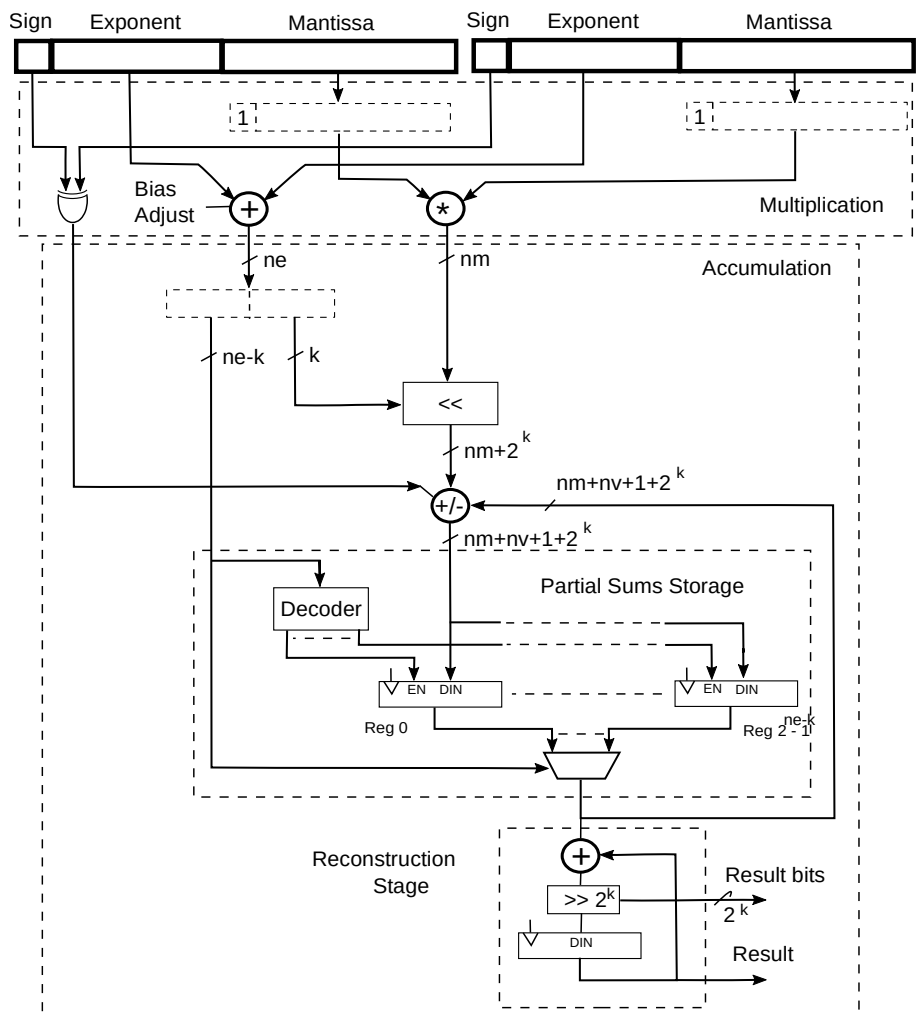

Figure 4 A multiply accumulator based on the described architecture.

Note that the circuit above works if we mix fixed point and floating point numbers. In fact, a fixed point number is just a number scaled by a fixed power of 2. Thus we just need to use such fixed power of 2 as exponent.

# 3. FPGA

The architecture described can be implemented particularly efficiently in AMD FPGAs. In fact, looking at the Partial Sums Storage in Fig.1,2 and 4, we realise that is identical to their distributed memory. In AMD FPGAs a single port 64x1 distributed memory requires a single LUT. This means that the partial sums storage, depending on nm and nv, can be implemented with a few tens of LUTs.

Since such storage constitutes the most resources, this results in a particularly efficient implementation. It is also possible to share the adder for the reconstruction stage, as mentioned multiple times.

Some examples of MAC for various floating point formats are shown in Tab.1, nv was set to 12. Results are with the fastest speed grade. These MACs perform a multiplication and an addition every clock cycle.

For fp8, the multiplier is constructed with directly with LUTs. These design use $k = 0$ (see section 2.2.).

For bfloat16, a DSP48E2 is used instead. The adder of the DSP48E2 primitive is also used. The design uses $k = 3$.

| Family | Kintex Ultrascale+ | | | Artix Ultrascale+ | | |
|---|---|---|---|---|---|---|
| fp format | LUTs | DSP48 | Max Freq (MHz) | LUTs | DSP48 | Max Freq (MHz) |
| fp8 E4M3 | 75 | 0 | ~630 | 75 | 0 | ~630 |
| fp8 E5M2 | 82 | 0 | ~740 | 85 | 0 | ~680 |
| bfloat16 | 97 | 1 | ~730 | 98 | 1 | ~630 |

Table 1 Implementation figures for AMD FPGAs.

These are minimalist implementations to demonstrate the technology. This means that, for example, minimum and maximum exponents are not tracked, resulting in longer clearing time of the partial sum registers.

## 3.1. A Tensor Core Example

Tensor cores are 4x4 matrices multiply accumulators that are part of the more modern NVIDA GPU architecture, used for speeding up matrix multiplications [2]. They perform the operation:

$$C_n = A_{n-1} B_{n-1} + C_{n-1} \quad \text{Eq.2}$$

Where A, B and C are 4x4 matrices. Using the bfloat16 MAC described above, we can now show how to build a tensor core that, every clock cycle, performs a matrix multiplication between two 4x4 bfloat16 matrices and accumulates the result with hundreds of bits of precision.

Expanding the recurrence in Eq.2, with the initial matrix $C_0$ being 0:

$$C_n = A_0 B_0 + \ldots + A_{n-1} B_{n-1} = \sum_{i=0}^{n-1} A_i B_i = \sum_{i=0}^{n-1} \begin{pmatrix} a_{i00} & \ldots & a_{i03} \\ \vdots & \ldots & \vdots \\ a_{i30} & \ldots & a_{i33} \end{pmatrix} \begin{pmatrix} b_{i00} & \ldots & b_{i03} \\ \vdots & \ldots & \vdots \\ b_{i30} & \ldots & b_{i33} \end{pmatrix} = \begin{pmatrix} c_{n00} & \ldots & c_{n03} \\ \vdots & \ldots & \vdots \\ c_{n30} & \ldots & c_{n33} \end{pmatrix} \quad \text{Eq.3}$$

With each element $c_{nrl}$ on row r and column l:

$$c_{nrl} = \sum_{i=0}^{n-1} \sum_{j=0}^{3} a_{irj} b_{ijl} = \sum_{j=0}^{3} \sum_{i=0}^{n-1} a_{irj} b_{ijl} = \sum_{i=0}^{n-1} a_{ir0} b_{i0l} + \sum_{i=0}^{n-1} a_{ir1} b_{i1l} + \sum_{i=0}^{n-1} a_{ir2} b_{i2l} + \sum_{i=0}^{n-1} a_{ir3} b_{i3l} \quad \text{Eq.4}$$

Looking at the last term of Eq.4, we notice four elements that are added together. Each of these represents a bfloat16 MAC operation repeated n times. If we want to perform a matrix multiplication and accumulation every clock cycle, we need 4 MACs for each of the elements of 4x4 matrices for a total of 64 bfloat16 MACs.

During the accumulation phase, each MAC will operate independently on the elements of the n incoming 4x4 matrices. After the accumulation phase we need the reconstruction phase but we also need to add the groups of 4 results as shown in the last term of Eq.4.

Fortunately these two operations can be combined with hardly any extra hardware. In fact, the DSP48E2 primitive that is used in the bfloat16 MAC allows multiple elements to be chained together. This chaining function allows the content of the accumulator of a DSP48E2 to be added to the one of the previous DPS48E2 element and pass it onto the next one.

All we need to do is to chain the 4 MACs that compute each of the last terms of Eq.4. During the reconstruction phase, each of 4 partial sums belonging to the same group of exponents (see section 2.2.) are added together by flowing through the chain of DSP48E2 before the add and shift in the final element of the chain.

Since the data flows to the chain continuously and without interruptions, apart for an extra 4 cycles latency, reconstruction takes exactly the same number of cycles as a single MAC.

The main problem is, however, how to keep feeding 32 bfloat16 numbers to such machine every clock cycle. Compression and quantization can help, see [3].

Another option is to calculate the first term of Eq.4 directly, using 16 bfloat16 MACs: this will require 4 clock cycle for each matrix multiplication instead of one.

# 4. ASIC

In this section, a rough gate (two input NAND) count estimation is given, with some consequent observations.

The assumption here is that the partial sum storage is made of flip-flops and the rest of ordinary logic. Also, since the reconstruction stage can be performed by re-using the adder and the existing registers, this will not be counted. Some other minor elements will also be neglected.

For the estimation, Appendix B of [4] was mostly used , with some differences. For example, for edge triggered D flip-flops, [4] uses 6 gates. Enabled D flip-flops are more appropriate in general as they allow to stall the design, so 9 gates were used instead. Also 5 and 9 gates wer used for half and full adder respectively, from [5].

Fig.2 contains the elements that were accounted:

- The barrel shifter that shifts the reconstructed mantissa by $0...2^k-1$ positions
- The adder/subtractor
- The $2^{ne-k}$ registers storing the partial sums
- The multiplexer that reads the $2^{ne-k}$ registers

Tab.2 shows the estimated gate count for various floating point formats and values of k from 0 to ne included. The number of bits nv stopping the partial sums from overflowing was arbitrarily set to 12.

| Format | ne | nm | k=0 | 1 | 2 | 3 | 4 | 5 | 6 | 7 | 8 |
|---|---|---|---|---|---|---|---|---|---|---|---|
| fp32 | 8 | 23 | 113,976 | 58,753 | 31,185 | 17,455 | 10,655 | 7,387 | 5,949 | 5,599 | 6,192 |
| bfloat16 | 8 | 7 | 64,776 | 34,081 | 18,753 | 11,119 | 7,353 | 5,563 | 4,845 | 4,831 | 5,505 |
| fp16 | 5 | 10 | 9,489 | 5,107 | 2,940 | 1,891 | 1,422 | 1,285 | - | - | - |
| fp8 E5M2 | 5 | 2 | 6,393 | 3,523 | 2,100 | 1,411 | 1,110 | 1,045 | - | - | - |
| fp8 E4M3 | 4 | 3 | 3,516 | 1,993 | 1,245 | 895 | 765 | - | - | - | - |
| fp8 E3M4 | 3 | 4 | 1,983 | 1,183 | 790 | 631 | - | - | - | - | - |

Table 2 Gate count estimation for some floating point formats.

As previously mentioned, when k = ne, we have the Kulisch accumulator. This seems to require the smallest number of gates in most cases. A couple of observations:

- Especially for larger fp formats, for the Kulisch case, the propagation delay through the barrel shifter and the add/sub is pretty long. This delay is progressively reduced as k gets smaller. It is possible that a synthesis tool, while attempting to achieve timing closure, might use a large number of resources, making the estimate reported here too optimistic for higher k values
- As k gets smaller, the number of outputs of the barrel shifter, add/sub and each register also gets smaller. Also, only one register at the time is active for each clock cycle. This means that the number of outputs that can potentially flip at each clock cycle is also reduced. This, in turn, has potential implications for the amount of power required by the circuit. Tab.3 shows the number of the outputs of the items listed in the bullet points above whose bits can flip at each clock cycle. This can be considered a crude proxy for power consumption.

For intermediate values of k, the data presented here hints at 2-4 times less output switching than the Kulisch accumulator: a possible sweet spot for gate count and power consumption. Power considerations are particularly important for some applications, such as AI processors, where hundreds, if not thousands of these circuits might be used.

| Format | ne | nm | k=0 | 1 | 2 | 3 | 4 | 5 | 6 | 7 | 8 |
|---|---|---|---|---|---|---|---|---|---|---|---|
| fp32 | 8 | 23 | 137 | 141 | 149 | 165 | 197 | 261 | 389 | 645 | 1,157 |
| bfloat16 | 8 | 7 | 73 | 77 | 85 | 101 | 133 | 197 | 325 | 581 | 1,093 |
| fp16 | 5 | 10 | 85 | 89 | 97 | 113 | 145 | 209 | - | - | - |
| fp8 E5M2 | 5 | 2 | 53 | 57 | 65 | 81 | 113 | 177 | - | - | - |
| fp8 E4M3 | 4 | 3 | 57 | 61 | 69 | 85 | 117 | - | - | - | - |
| fp8 E3M4 | 3 | 4 | 61 | 65 | 73 | 89 | - | - | - | - | - |

Table 3 Number of outputs that can potentially flip at each clock cycle.

This analysis is based on the partial sums storage area implemented as flip-flops but register file implemented with RAM cells is another possibility although that would require tests with specific target technologies.

## 4.1. A Parallel Implementation

There are some cases we need to combine the result of multiple MAC operations. As we saw in the tensor core implementation, we can have multiple MACs operating independently, only combining their results during the reconstruction phase.

However, when doing that, each MAC has its own copy of the partial sum registers. This is not very efficient as they ultimately need to be added together: it would be much better if we could add the incoming numbers directly into a single set of partial sum registers. It might not be worth doing in FPGAs since distributed memories are so efficient, but it should be for ASIC, especially since partial sum registers are such a large portion of the design.

Fig.5 shows a parallel adaptation of the architecture described in section 2.2. For simplicity, the case of adding 4 floating point numbers is illustrated. The objective here is to add four floating point numbers every clock cycle.

We obviously need a multi-port implementation of the partial sum registers and a mechanism to avoid simultaneous writing to the same location of the multi-port register file.

Four numbers arrive at the same time, the mantissas are reconstructed and converted to two's complement. Next, the mantissas are shifted, as shown previously. The results, together with their group exponent indexes, are forwarded to the Route & Add module.

The Route & Add module cannot be a simple non blocking switch because we could have multiple mantissas belonging to the same exponent group, causing multiple writes to the same location of the partial sum register file.

We need something that adds all such mantissas with the same exponent group together before presenting them to the multi-port version of the partial sum registers. Also, appropriate write enable signals (active high here) need to be generated. This information is then used to read and update the corresponding partial sums.

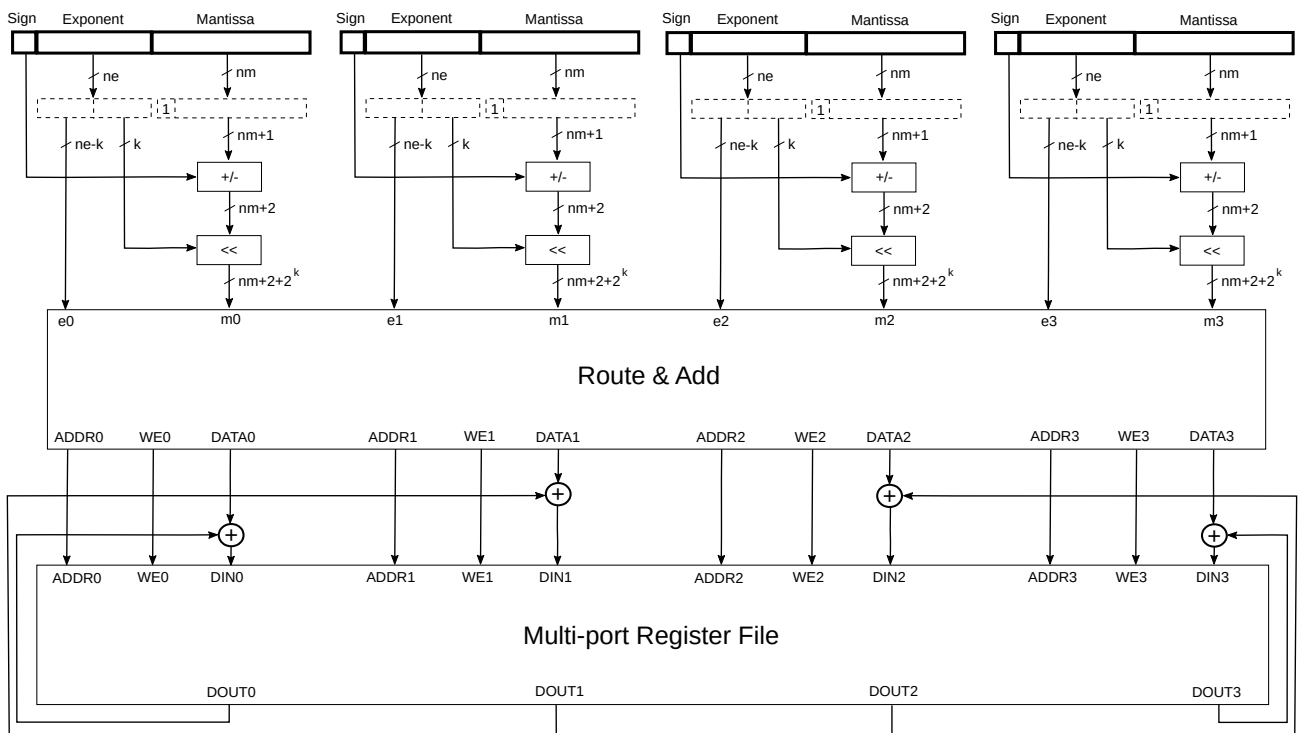

Figure 5 Parallel implementation of the exponent indexed accumulator.

Reconstruction is not shown in the figure because it's exactly as before and, as before, it can be implemented by sharing the adder and one of the registers with the accumulation stage. The design can also be pipelined.

Fig.6 shows a possible implementation of the Route & Add module. Each exponent group is used directly as an address for the register file, without further processing. Each input m0, m1, etc. has its own circuit that controls the addition to other inputs.

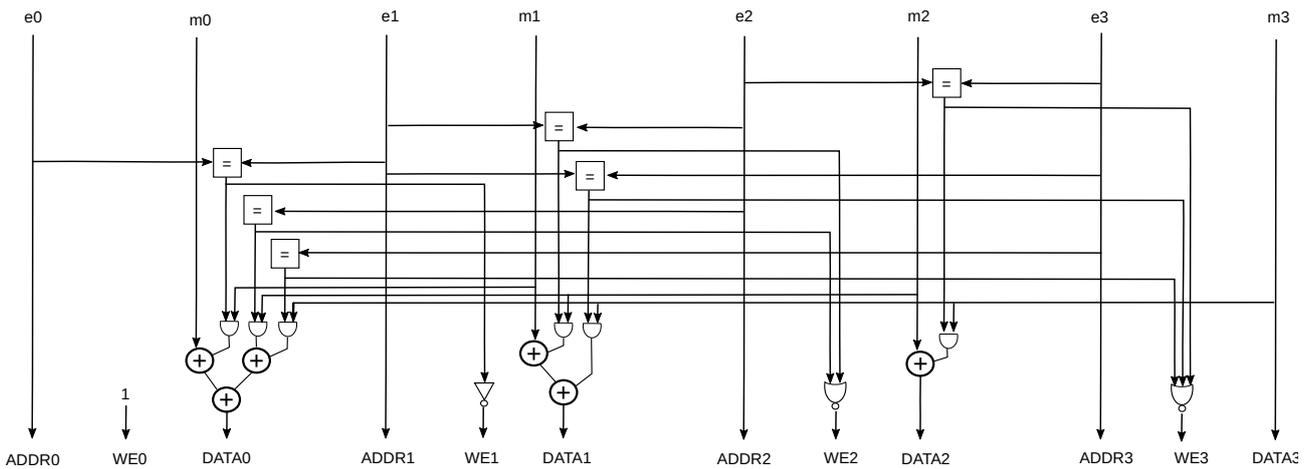

Figure 6 A possible implementation of the Route & Add module.

Looking from left to right, the mantissa m0 is added to m1, m2 and m3 but only if they belong to the same exponent group. The addition process is controlled by equality comparisons to the exponent groups. The result of the comparisons are used to enable the inputs of a tree adder.

The results of the very same comparisons is also used to disable the other write enables and avoid two or more simultaneous and incorrect writes to the same partial sum register.

A similar circuit is used when adding m1 to m2 and m3. Note that, in this case, we do not need to consider m0. In fact, if m0 and m1 share the same exponent group, they have already been added by the previous circuit on the left. And, if they are different, they belong to separate write operations to a different partial sum register. Either way, any input does not need to consider inputs to its left.

The circuit in Fig.4 and Fig.5 can be generalised from four to N floating point numbers. The N incoming floating point numbers are pre-processed as before We also need an N-port register file containing the partial sums.

For the Route & Add module, each input $m_i$ is added to $m_{i+1}$, $m_{i+2}$,.....$m_{N-1}$ but only if $e_i$ is equal to $e_{i+1}$ or $e_{i+2}$, ....,$e_{N-1}$. Each equality comparison controls the tree adder inputs and, as before, it is also used to disable other writes to same locations.

# 5. MACs for Posits

Just a quick note on posits [6]. Essentially, posits can be seen as being equivalent to floating point numbers with a variable size mantissa whose size depends on the value of the exponent. So, they can still be converted to and from a $m2^e$ form, it's just that the number of bits for m is variable.

We can still use the architectures described in section 2. but the partial sum registers will have a variable number of bits. During the reconstruction phase they will need to be padded with zeros so that they can be added together.

# 6. Logarithmic MACs

Logarithmic number representation has recently emerged as an alternative to floats in applications where low number of bits are possible such as neural networks. Given the same number bits, logarithmic numbers have a better accuracy than floats [7]. Such numbers are represented as:

$$v=(-1)^s 2^{ei.ef}=(-1)^s 2^{ei+0.ef}=(-1)^s 2^{ei} 2^{0.ef}=(-1)^s m 2^{ei} \quad \text{Eq.5}$$

Where ei and ef are, respectively, the integer and the fractional part of the exponent, with $m=2^{0.ef}$. The format looks like this:

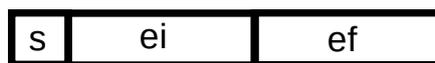

Figure 7 Logarithmic number format

Effectively, we have an unsigned fixed point number and a sign. Note that zero cannot be represented directly: we can either use the smallest number that can be represented instead or dedicate a special code to it (i.e., unimaginatively, 0).

This representation is particularly convenient because multiplications can be done with additions. Unfortunately, things get more complicated when adding numbers in logarithmic form as they need to be converted to ordinary numbers before being actually added. This, in turn, can result in loss of precision, especially over a large number of additions.

However, given the very high precision of architecture described, this problem is substantially reduced. Yes, the fractional part of the logarithmic representation is necessarily converted to a finite number of digits but, after that, there's no more loss of accuracy, subject to the chosen reconstruction strategy (section 2.3.).

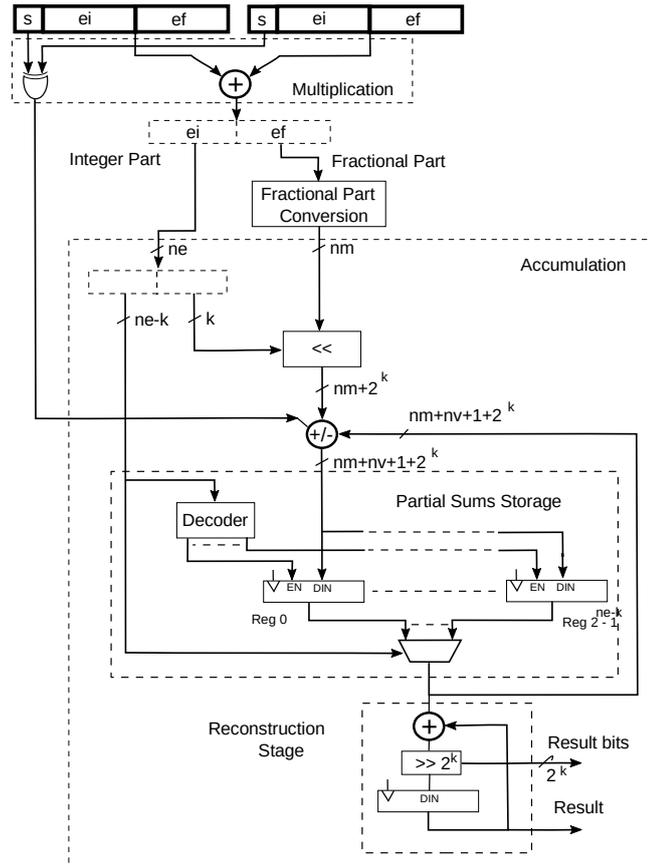

Figure 8 Logarithmic numbers multiply accumulator.

Fig.8 shows a possible implementation of a logarithmic MAC. For the multiplication, the sign of the result is obtained by xoring the signs of the two operands while the result itself is obtained by simply adding the two fixed point numbers. The result will also have an integer and fractional part. As the resulting fractional part is generally only a few bits, it can be converted to $2^{0.ef}$ with a simple lookup table.

After this, we have a sign, an exponent and a mantissa that can be then accumulated as previously described. All the previously considerations made about this architecture are also valid here.

## 6.1. FPGA Implementation

The architecture described above was implemented in AMD FPGAs for 8 bits logarithmic numbers with 4 bits integer and 3 bits fractional part (log4.3 format). The design is very similar to the fp8 designs shown in section  3.. The main difference is that, instead of a multiplier, we have an adder and the conversion lookup table. The parameter k = 0 was chosen (see section  2.2.), nv was set to 12. Results are with the fastest speed grade. This MAC perform a multiplication and an addition every clock cycle.

In this particular implementation the conversion lookup table maps the 3 bits of the fractional part into 8 bits numbers representing $2^{0.ef}$. This is the same accuracy as bfloat16 which is more than adequate for many applications. In any case this is easy to change.

| Kintex Ultrascale+ | | | Artix Ultrascale+ | | |
|---|---|---|---|---|---|
| LUTs | DSP48 | Max Freq (MHz) | LUTs | DSP48 | Max Freq (MHz) |
| 74 | 0 | ~618 | 75 | 0 | ~618 |

Table 4 Implementation figures for AMD FPGAs.

## 6.2. Logarithmic Numbers Compression

This is a bit off topic but it's a good opportunity to check how well logarithmic numbers would compress within the framework presented in [3].

For this purpose, the LLama2 7B weights were used, same as in [3]. Unfortunately the LLama2 7B weights would require at least 5 bits for the integer exponent and so they do not fit in the log4.3 format directly.

Thus they were linearly scaled to 16 bits + sign first. Each of the weight matrices $W_{ij}$ were scaled separately with:

$$\frac{2^{16}-1}{max(|(W_{ij})|)} W_{ij} \quad \text{Eq.6}$$

Now the integer part of base 2 logarithm of each value fits in the 4 bits of the log4.3 format and can be cast to it. A special code was created for the value zero.

Next, for each weight matrix separately, each value was encoded with coding pairs. Two different methods were used, shown in Fig.9.

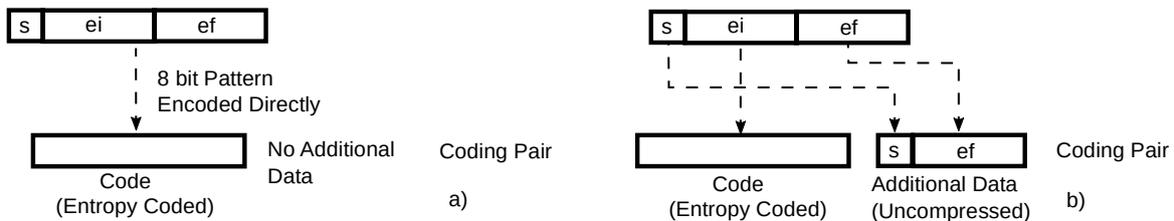

Figure 9 Logarithmic numbers compression as coding pairs.

In the first case, logarithmic numbers are entropy coded directly with no additional data (Fig.9a). In the second case (Fig.9b), only the integer part ei is entropy coded, with the sign and ef part of the additional data. A special code for zero, with no additional data is also used.

| Method | File size (bytes) | % of original size | Avg weight (bits) |
|---|---|---|---|
| Log4.3 | 6,607077,085 | 100 | 8 |
| a) | 5,389,527,617 | ~81.56 | ~6.526 |
| b) | 5,428,798,248 | ~82.17 | ~6.573 |

Table 5 A comparison of the methods a) and b).

The original bfloat16 file used in [3] is 13,214,154,752 bytes. When converted to log4.3 format, it is halved to 6,607,077,085 bytes (from 16 bits of bfloat16 to 8 of log4.3). This is shown in Tab.5.

Tab.5 also shows the size of the weights in log4.3 format when losslessly compressed as shown in Fig.9a and Fig.9b. This is only an estimate obtained estimating the probability of a code p and using $-\log_2(p)$ as the number of bits required. However, as shown in [3], it's easy to get close to the theoretical limit with simple hardware.

Note that both methods produce a very similar result. However, the second method requires a much smaller and simpler probability model (17 vs. 256 codes). This will be reflected in the simplicity of the implementation.

# 7. Conclusion

A fast and simple method to add floating point, posits and logarithmic numbers was presented.

The implementation is particularly efficient in FPGAs where dealing with floating point numbers is resource consuming.

Although more work is needed to confirm it, these results imply a potential reduction in power in ASIC applications, possibly at a cost of a somewhat larger gate count.

The potential exact calculation of the result should also benefit scientific applications that are affected by accumulating rounding errors.

Finally, the method might be potentially used in software with processors without a floating point unit.

## References

[1] U. Kulisch. Computer arithmetic and validity: Theory,implementation, and applications. 2013.
[2] NVIDIA Docs Hub, "Matrix Multiplication Background User's Guide", https://docs.nvidia.com/deeplearning/performance/dl-performance-matrix-multiplication/index.html
[3] V. Liguori(2024). From a Lossless (~1.5:1) Compression Algorithm for Llama2 7B Weights to Variable Precision, Variable Range, Compressed Numeric Data Types for CNNs and LLMs .arXiv preprint arXiv:2404.10896
[4] Mart van Baalen,…,Tijmen Blankevoort (2023) FP8 versus INT8 for efficient deep learning inference. arXiv preprint arXiv:2303.17951
[5] Wikipedia. Adder. https://en.wikipedia.org/wiki/Adder_(electronics)
[6] John Gustafson, "Beating Floats at Their Own Game ", https://www.youtube.com/watch?v=N05yYbUZMSQ
[7] Bill Dally (NVIDIA), "Trends in Deep Learning Hardware", https://www.youtube.com/watch?v=kLiwvnr4L80